# Reasonable Anomaly Detection in Long Sequences

Yalong Jiang, *Member*, *IEEE*, Changkang Li

**Abstract**

Video anomaly detection is a challenging task due to the lack in approaches for representing samples. The visual representations of most existing approaches are limited by short-term sequences of observations which cannot provide enough clues for achieving reasonable detections. In this paper, we propose to completely represent the motion patterns of objects by learning from long-term sequences. Firstly, a Stacked State Machine (SSM) model is proposed to represent the temporal dependencies which are consistent across long-range observations. Then SSM model functions in predicting future states based on past ones, the divergence between the predictions with inherent normal patterns and observed ones determines anomalies which violate normal motion patterns. Extensive experiments are carried out to evaluate the proposed approach on the dataset and existing ones. Improvements over state-of-the-art methods can be observed. Our code is available at https://github.com/AllenYLJiang/Anomaly-Detection-in-Sequences.

## 1. Introduction

Due to the large variety of abnormal events and inaccessibility of task-specific data, anomaly detection is a quite challenging research problem in surveillance. Typical anomalies include robbing, burglary and so on. However, abnormal actions rarely occur and do not conform to any fixed pattern, it is difficult to obtain annotations on anomalies. As a result, unsupervised approaches are required to distinguish the events that do not match regular patterns [1] [2].

Existing unsupervised or weakly supervised approaches leverage either the unpredictability of human behaviors [3] [4] [5] [6] [7] [8] [9] [10] [11] [12] or the divergence in deep features [13] [14] [15] [16] [17] [18] [19] [20] between normal and abnormal events. To address the lack in data, some other approaches [21] such as [22] [23] [24] have also been proposed to explore the separability of motion patterns. Furthermore, [25] [26] [27] proposed to leverage the characteristics of abnormal actions in separating irregular behaviors from normal ones.

Although impressive results have been achieved, the above approaches have not considered whether anomaly detections are reasonable. For instance, short-term sequences of observations cannot well describe the properties about actions [14] [4]. Some short-term motion patterns are shared by regular and irregular behaviors, it's difficult to characterize actions with short sequences. Besides, short-term observation noises [28] lead to incorrect and counterfactual pose estimations which result in implausible anomaly detections, the detections violate the long-term consistent patterns governing human motion. As a result, describing human behaviors with long-term motion representations is necessary in achieving reasonable anomaly detections. In one way, the effective encoding of motion patterns underlying long sequences provides comprehensive clues for distinguishing actions while exhibiting robustness to short-term random variations. On the other way, short-term implausible motion estimations can be filtered out with long-term consistent patterns.

Furthermore, some of the methods can only conduct frame-level anomaly detection without localizing abnormal regions [29] [7], the interpretability of these methods is not satisfactory because the underlying factors leading to anomaly may come from background clutters. As a result, subject-level anomaly detection facilitates the explorations of individuals' dynamics and is less influenced by backgrounds.

Targeted at achieving complete and reasonable representations about objects' movements, an approach is proposed to effectively learn the motion patterns from long sequences of observations.

The contributions of this paper are summarized as follows: SSM models based on state machines are proposed for completely representing the motion patterns underlying long-term sequences of actions. The representations are leveraged in distinguishing anomalies through predicting future states based on past observations. Extensive experiments are conducted to demonstrate the superiority of the proposed method on existing datasets.

## 2. Related Works

Due to the imbalance of surveillance videos, unpredictable nature of anomalies and inaccessibility of annotations, existing approaches are mostly unsupervised. Reconstruction or prediction-based approaches produce larger error on irregular motion patterns than on normal ones. For instance, [13] proposed an encoder-decoder based network for reconstruction with inherent probabilistic modeling on latent feature encodings. [14] introduced attention mechanism into encoder-decoder structures and evaluated the correlations between inputs and different memory elements for querying. [15] and [16] combined LSTM structures with encoder-decoder networks for fall detection. [17] proposed to integrate appearance with motion-related clues as inputs to encoders. [18] compressed each video into a single frame before augmenting reconstruction loss with sparsity constraints. Other reconstruction-based methods include autoencoders [30] and adversarially learned models [31]. Prediction-based methods evaluate the divergence between the motion patterns in later frames and those in past ones. Typical ones include skeleton prediction [4] [5], LSTM-based prediction [6], GAN-based prediction [7], variational autoencoder-based prediction [8] and spatio-temporal two stream predicting models [9]. [10] [11] [19] [12] proposed to combine prediction with reconstruction and built a pool of prototype features for encoding normal dynamics, a few frames were also leveraged to fine-tune hyperparameters to adapt to new scenes. Although remarkable improvements have been achieved [4], most of the above-mentioned approaches are based on the parsing of short sequences which cannot completely describe objects' behaviors. Besides, short-term observations are easily influenced by noises such as occlusions, producing implausible results. As a result, we move toward comprehensive understandings about objects' behaviors and plausible anomaly detections by effectively encoding long-term motion patterns.

Aside from the above-mentioned approaches which only learn from normal data, distance-based approaches built similarity metrics between video instances. For instance, clustering-based approaches measured the similarity in humans' spatial and temporal embeddings [22] [23] [32] [33], [24] proposed to use 3D convolutions in building spatial and temporal representations. To improve the measurement of similarity in clustering, [34] enlarged the distances between normal and abnormal events while ensuring inner-class compactness, however, it required the labels on anomalies. [32] proposed to enlarge inter-class distances to better distinguish anomalies. [35] proposed to model anomaly detection as a one-versus-rest classification task. To build representations in complex scenarios, [21] proposed a probabilistic framework for categorizing actions. [36] proposed to build a graph connecting different objects in each frame and cluster graphs. [27] was targeted at crowded scenes and proposed to integrate collective properties for multi-stage clustering. Differently, we propose to build complete representations to ensure valid distance computations and reasonable outlier detections.

To generalize to novel circumstances, meta learning-based methods such as [11] [12] introduced adjustable feature representations which can adapt to new domains. Attention-based methods such as [37] were proposed to attend to critical and domain-invariant features while reducing the influence of backgrounds. [38] conducted sparse encoding to focus on invariant motion features. To further improve generalization by combining complementary task-specific clues, [39] integrated multiple sub-tasks, including moving direction prediction, appearance consistency evaluation and object classification to better align with anomaly detection. [40] proposed an approach based on multi-instance learning (MIL). [26] introduced casual temporal relations to enhance MIL-based approaches and developed compact and discriminative representations with causal temporal relations. [41] [42] applied robust representations in addressing unexpected feature patterns. [43] introduced invariant rules in achieving robust anomaly detections. Different from [43], our proposed approach deals with high-dimensional video data with time-variant probability distributions. An approach based on state machines is proposed to obtain the consistent motion patterns that generalize to long-term temporal variations. As is shown by [44], state machines are better at modeling the intrinsic motion patterns in long sequences than other methods such as transformers [45], exhibiting robustness to both short-term domain-shifts and observation noises.

## 3. Method

In this section, we propose a SSM (Stacked State Machine) model which encodes the relations between objects' states and the variations in states across long periods. The contributions are in two folds. Firstly, SSM model extracts the motion patterns which are consistent across long sequences. The patterns are encoded by state machines which predict future states based on past ones, the divergence between predictions and future observations is leveraged in determining anomalies, as will be detailed in Section 3-A and B.

*A. Modeling Long-Range Temporal Dependencies in Actions with State Machines*

Different from short-range temporal dependencies which vary across the phases of actions or random changes in view-points, long-range dependencies provide more comprehensive clues in representing actions. In this section, state machines are leveraged in discovering the intrinsic motion patterns governing objects' long-range movements and characterizing objects' actions with the patterns.

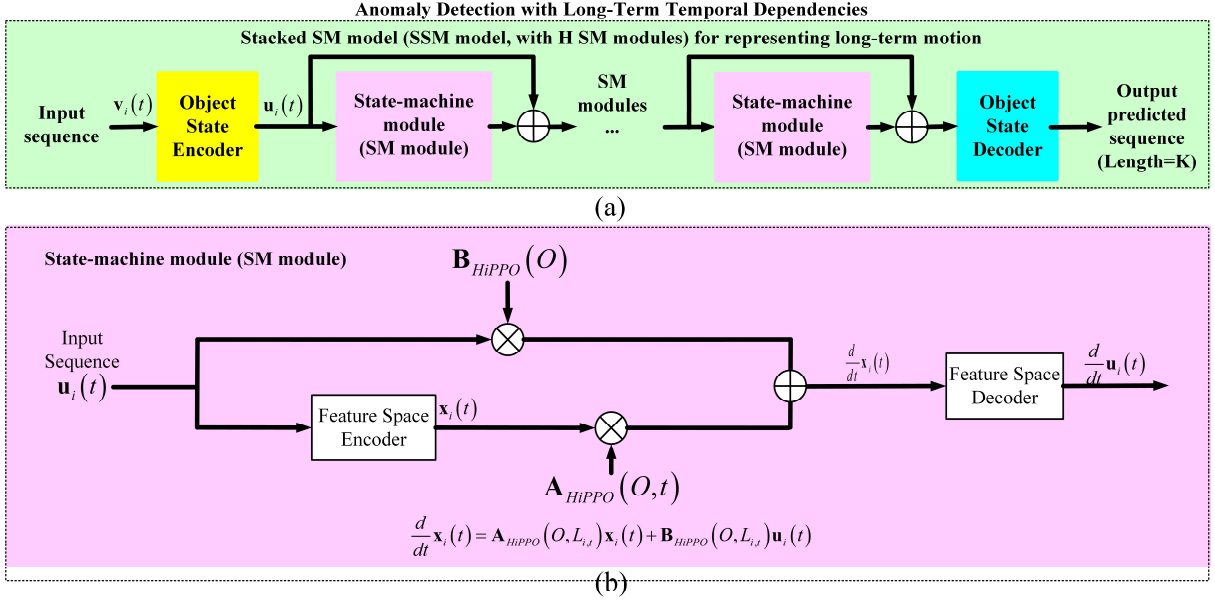

Fig. 1. Illustration of the proposed approach

**The structure of SSM model.** State machines are used in modeling the temporal dependencies between past and future states. SSM model is shown by Fig. 1(a), it stacks $H$ SM modules each of which is illustrated in Fig. 1(b). The dependencies are encoded with weights that are not data-driven, avoiding motion representations from being over-fitted to limited training patterns. Besides, the modules can adapt to the inputs with variable durations under incomplete observations.

Denote $\mathbf{v}_i(t) \in \mathbb{R}^{N \times L_{i,t}}, \forall i$ the states (poses and locations) of human $i$ from $t_i$ to $t$, as is shown by Fig. 1(a). $t_i = t - L_{i,t} + 1$ is the first moment when human $i$ is detected. $N$ has value 34, representing the 2-dimensional coordinates of 17 body joints. SSM model firstly encodes $\mathbf{v}_i(t)$ into $\mathbf{u}_i(t) \in \mathbb{R}^{O \times L_{i,t}}, O = 64$. Each SM module takes $\mathbf{u}_i(t)$ as input and outputs its time derivative $d\mathbf{u}_i(t)/dt$. Specifically, $\mathbf{u}_i(t)$ is encoded with $\mathbf{x}_i(t) \in \mathbb{R}^{O \times L_{i,t}}$. Each row in $\mathbf{x}_i(t)$ shows one part of human $i$, each column in $\mathbf{x}_i(t)$ corresponds to one moment. The encoders and decoders in Fig. 1(a) and (b) model the spatial relations between human $i$'s parts. The function of SM module is shown by a differential equation in Eq. (1):

$$\frac{d}{dt}\mathbf{x}_i(t) = \mathbf{A}_{HiPPO,i}(O, L_{i,t})\mathbf{x}_i(t) + \mathbf{B}_{HiPPO,i}(O, L_{i,t})\mathbf{u}_i(t) \tag{1}$$

Different from transformers and LSTMs [46] [19] where the temporal relations between past and futures states are learned from data, the $\mathbf{A}_{HiPPO,i}(O, L_{i,t})$ and $\mathbf{B}_{HiPPO,i}(O, L_{i,t})$ are not data-driven. Due to the random variations in view-points and scales, the learned temporal dependencies between past and future states cannot generalize, this is the case in both normal and abnormal actions. The entries in $\mathbf{A}_{HiPPO,i}(O, L_{i,t}) \in \mathbb{R}^{O \times L_{i,t}}$ and $\mathbf{B}_{HiPPO,i}(O, L_{i,t}) \in \mathbb{R}^{O \times L_{i,t}}$ are set according to [44], facilitating the memorization of long-term dependencies by extracting frequency components:

$$\mathbf{A}_{HiPPO,i}(o,l) = -\begin{cases} \left((2o+1)^{1/2}(2l+1)^{1/2}\right)/L_{i,t}, & o > l \\ (o+1)/L_{i,t}, & o = l, o \in [1,O], l \in [1, L_{i,t}] \\ 0, & o < l \end{cases} \tag{2}$$

$$\mathbf{B}_{HiPPO,i}(o,l) = (2o+1)^{1/2}, o \in [1,O], l \in [1, L_{i,t}] \tag{3}$$

For fixed $t_i$, $L_{i,t}$ grows as $t$ increases, allowing more observations to be leveraged in determining $d\mathbf{x}_i(t)/dt$.

The reason for stacking $H$ SM modules instead of one in SSM model is to achieve a better discretization. Specifically, each state machine only needs to predict the variations in states across a period with length $\Delta = K/H$, achieving

$$\delta \mathbf{x}_i(\tau)\bigg|_{\tau=t} = \frac{d}{d\tau}\mathbf{x}_i(\tau)\bigg|_{\tau=t} \cdot \Delta \tag{4}$$

$\delta \mathbf{x}_i(t)$ can better approximate the variations in $\mathbf{x}_i(t)$ for smaller $\Delta$. $H$ is 3 in our implementations. The hyper-parameters of the encoders and decoders in Fig. 1(a) and (b) are shown by Table I:

TABLE I
HYPER-PARAMETERS OF ENCODERS AND DECODERS IN SSM MODEL, MLP: MULTI-LAYER PERCEPTRON

| Module name | Module type | Hyperparameters | |
|---|---|---|---|
| | | Number of input channels | Number of output channels |
| Object State Encoder | MLP | 34 | 64 |
| Object State Decoder | MLP | 64 | 34 |
| Feature Space Encoder | MLP | 64 | 64 |
| Feature Space Decoder | MLP | 64 | 64 |

*B. SSM Model for Anomaly Detection*

The inputs to SSM model are provided by detector and pose estimator. We model SSM model's function in Eq. (5):

$$SSM_2\big(\mathbf{v}_i(t)w(t,L_{i,t})\big) = \hat{\mathbf{v}}_i(t,t+1),...,\hat{\mathbf{v}}_i(t,t+K) \tag{5}$$

where $\mathbf{v}_i(t) = \mathbf{v}_i(t,1),...,\mathbf{v}_i(t,t)$ denotes all past states in human $i$'s trajectory, $w(t,L_{i,t})$ is a temporal window with size $L_{i,t}, L_{i,t} \leq t$, keeping the states within period $[t-L_{i,t}+1,t]$ while discarding earlier ones, $\mathbf{v}_i(t)w(t,L_{i,t}) = \mathbf{v}_i(t,t-L_{i,t}+1),...,\mathbf{v}_i(t,t)$. The anomaly score of human $i$ at $t+1$ is obtained with the sum of divergence between $\hat{\mathbf{v}}_i(t,t+k)$ and $\mathbf{v}_i(t,t+k)$ for all $k$:

$$s_i(t+1) = \sum_{k=1}^{K} MSE\big(\hat{\mathbf{v}}_i(t,t+k), \mathbf{v}_i(t,t+k)\big) \tag{11}$$

where $MSE()$ denotes mean square error. SSM model only encodes the motion patterns in training data, the state machines in SSM cannot generate realistic pose sequences when observing abnormal input motion patterns, leading to higher prediction error on anomalies.

Each score $s_i(\tau), \forall \tau$ is normalized with respect to the number of body joints because $N$ is 34 for humans and is 8 for non-human objects. In each frame, the anomaly score is the maximum one among all objects:

$$s(t) = \max_i s_i(t) \tag{12}$$

4. **Experiments**

*A. Introduction to Datasets*

The proposed approach is also evaluated on existing benchmarks: the CUHK Avenue dataset [47] and ShanghaiTech [38]. There are two versions of ShanghaiTech dataset [38] [48], the latter contains the annotations of anomalies in training set. As our approach is targeted at practical applications where no anomaly labels are provided, we use the first version.

ROC (Receiver Operation Characteristic) Curve (AUC) is leveraged for evaluation [7] [17]. The metric is computed by continuously changing the threshold of anomaly scores. Then AUC is obtained through cumulating under the ROC curve. A higher AUC value indicates a better performance. We evaluate the performance with two clearly defined frame-level AUCs as metrics: Macro-averaged AUC (macro-AUC) first computes the AUC for each video, then averages the resulting AUCs of videos, and micro-averaged AUC (micro-AUC) first concatenates the scores of all videos and then computes AUC.

*B. Implementation Details*

We adopt Yolo [49] for detecting objects. It is pretrained on COCO [50]. In the computation of anomaly scores, temporally sliding windows with stride 1 are adopted. For human $i$, the length of sliding window for determining the anomaly score at moment $t+1$ is $L_{i,t} + K$, as is shown in Eq. (11). To reduce computational complexity, an upper limit on window size is set, as will be shown in ablation studies. The algorithm runs on an Intel Core i7 processor and NVIDIA RTX-3080 GPU.

The training set includes each human sequence with length $L_{i,t} + K$ at $t$. The input tensor has shape $B \times (L_{i,t} + K) \times N$ where $B = 256$ denotes batch size. The pre-training lasts for 15 epochs with Adam optimizer and learning rate 5e-5. The learning rate decay parameter is 0.99, learning rate is multiplied by 0.99 after each epoch. The SSM model is frozen after training.

TABLE II
FRAME-LEVEL AREA UNDER THE CURVE COMPARISON ON EXISTING BENCHMARK DATASETS

| | Algorithm | ShanghaiTech | | CUHK Avenue | |
|---|---|---|---|---|---|
| Reconstruction or prediction | Normal Dynamics [10] | 73.8 | - | 89.5 | - |
| | Normal Graph [5] | 74.1 | - | 87.3 | - |
| | Memorizing Normality [14] | 72.8 | - | 84.9 | - |
| | Frame Prediction [7] | 72.8 | - | 85.1 | - |
| | Deepfall [16] | - | - | - | - |
| | Object-centric [35] | 84.9 | - | 90.4 | - |
| | Jigsaw Puzzles [51] | 84.3 | - | 92.2 | - |
| | Learning Regularity [4] | 73.4 | - | - | - |
| | Stacked RNN [38] | - | - | 81.71 | - |
| | Old is Gold [52] | - | - | - | - |
| Transfer / Meta | Latent Space [13] | 72.5 | - | - | - |
| | Continual Learning [53] | 71.62 | - | 86.4 | - |
| | Anomaly3D [24] | 80.6 | - | 89.2 | - |
| | Few-shot [11] | 77.9 | - | 85.8 | - |
| | Memory Guided [12] | 70.5 | - | 88.5 | - |
| Clustering | Clustering Driven [17] | 73.3 | - | 86.0 | - |
| | Margin Learning [34] | 76.8 | - | 89.2 | - |
| | Scene-Aware [36] | 74.7 | - | 89.6 | - |
| | Graph-embedded [54] | 76.1 | - | - | - |
| | Local Aggregation [55] | 74.7 | - | 89.9 | - |
| Multi-task | Multi-Task [39] | - | 90.2 | - | 92.8 |
| | Self-supervised [56] | - | 89.5 | - | 92.9 |
| | Street Scene [57] | - | - | - | - |
| | AI-VAD [58] | 85.9 | 89.6 | **93.3** | **96.2** |
| | Ubnormal [59] | - | 90.5 | - | 93.2 |
| | Background-agnostic [60] | - | 89.3 | - | 92.3 |
| | AnomalyNet [18] | - | - | - | - |
| Ours | Ours without the relative offsets between objects in one sequence in Eq.(8) | 87.0 | 90.1 | 92.4 | 95.2 |
| | **Ours** | **88.8** | **91.8** | **93.3** | **96.1** |

## C. Quantitative Results

The proposed approach is compared with other methods on existing benchmarks, as is shown in Table II. The baselines include reconstruction-based or prediction-based approaches such as [4] [14], transfer-learning or meta-learning based approaches such as [53], clustering-based approaches such as [54], and multi-task approaches such as [39]. It can be observed that the proposed state machine-based approach outperforms existing ones. Specifically, the accuracy in Table II is achieved with $L_{i,t}, \forall i,t$ being truncated by an upper limit of 30. In the early moments when human $i$ just starts to appear, $L_{i,t} < 30$. Efficiency decreases if without truncation.

Besides, the comparison between the last two layers shows the advantage of coupling pose variations with the relative positions between observations of the same object at different moments. As is addressed in [61], velocities are indicators of anomalies, the coupling provides a more comprehensive representation about movements.

## 5. Conclusion

In this article, an approach for anomaly detection is proposed to conduct anomaly detection with long sequences of observations. A framework based on state machines is proposed for modeling the long-term dependencies which comprehensively describe human movements and thus ensure reasonable anomaly detections. The proposed approach outperforms state-of-the-art visual methods on both existing benchmarks and the proposed dataset. The approach generalizes to non-human objects, including vehicles, bicycles and so on.